\documentclass{article}

\usepackage{PRIMEarxiv}

\usepackage[utf8]{inputenc} 
\usepackage[T1]{fontenc}    
\usepackage{hyperref}       
\usepackage{url}            
\usepackage{booktabs}       
\usepackage{amsfonts}       
\usepackage{nicefrac}       
\usepackage{microtype}      
\usepackage{lipsum}
\usepackage{fancyhdr}       
\usepackage{graphicx}       
\graphicspath{{media/}}     
\usepackage{bm}
\usepackage{amsmath}
\usepackage{amssymb}
\usepackage{enumerate}    
\usepackage{amsfonts}
\usepackage{bbm} 
\usepackage{algorithmic}
\usepackage{algorithm}
\usepackage{array} 
\usepackage{textcomp}
\usepackage{stfloats}
\usepackage{url}
\usepackage{verbatim}
\usepackage{color}
\usepackage{subcaption}
\usepackage{graphicx} 
\usepackage{xcolor, soul}
\title{Interconnection and Damping Assignment Passivity-Based Control using Sparse Neural ODEs}

\author{
  Nicolò Botteghi \\
  Department of Mathematics \\
  Politecnico di Milano \\
  Milano, Italy \\
  \texttt{nicolo.botteghi@polimi.it} \\
   \And
  Owen Brook \\
  Department of Aeronautics \\
  Imperial College London \\
  London, United Kingdom\\
  \texttt{owen.brook20@imperial.ac.uk} \\
   \AND
  Urban Fasel \\
  Department of Aeronautics \\
  Imperial College London \\
  London, United Kingdom\\
  \texttt{u.fasel@imperial.ac.uk} \\
  \And
  Federico Califano \\
  Department of Robotics and Mechatronics\\
  University of Twente \\
  Enschede, Netherlands \\
  \texttt{f.califano@utwente.nl}\\
}

\begin{document}
\maketitle

\begin{abstract}
Interconnection and Damping Assignment Passivity-Based Control (IDA-PBC) is a nonlinear control technique that assigns a port-Hamiltonian (pH) structure to a controlled system using a state-feedback law. 
While IDA-PBC has been extensively studied and applied to many systems, its practical implementation often remains confined to academic examples and, almost exclusively, to stabilization tasks. The main limitation of IDA-PBC stems from the complexity of analytically solving a set of partial differential equations (PDEs), referred to as the \emph{matching conditions}, which enforce the pH structure of the closed-loop system. However, this is extremely challenging, especially for complex physical systems and tasks.

In this work, we propose a novel numerical approach for designing IDA-PBC controllers without solving the matching PDEs exactly. We cast the IDA-PBC problem as the learning of a \emph{neural ordinary differential equation}. In particular, we rely on \emph{sparse dictionary learning} to parametrize the desired closed-loop system as a sparse linear combination of nonlinear state-dependent functions. Optimization of the controller parameters is achieved by solving a \emph{multi-objective optimization} problem whose cost function is composed of a generic task-dependent cost and a matching condition-dependent cost. Our numerical results show that the proposed method enables \emph{(i)} IDA-PBC to be applicable to complex tasks beyond stabilization, such as the discovery of periodic oscillatory behaviors, \emph{(ii)} the derivation of closed-form expressions of the controlled system, including residual terms in case of approximate matching, and \emph{(iii)} stability analysis of the learned controller.
\end{abstract}


\section{Introduction}
Port-Hamiltonian (pH) systems are nonlinear affine systems encoding a rich set of system-theoretical and behavioral properties, conveniently modeling networks of physical systems in any physical domain \cite{Duindam2009ModelingSystems,Rashad2020TwentyReview}. PH systems describe the energy exchanges occurring within the modeled physical systems, such as energy routing, dissipation, and external energy transfers through e.g., control inputs or external systems. From a control-theoretic perspective, a pH system encodes desirable properties such as passivity, stability, and general energetic behaviors that can serve as performance metrics in complex tasks \cite{Ortega2001PuttingControl,ortega2002interconnection}. It is these properties that motivate control methods aiming to obtain a closed-loop pH structure. Among these energy-based techniques, \emph{Interconnection and Damping assignment Passivity-Based Control} (IDA-PBC) \cite{ortega2004interconnection,ortega2002interconnection} is one of the most popular, as it assigns a general pH structure to a controlled physical system using a nonlinear state-feedback control law.

Despite its popularity, IDA-PBC has remained largely limited to academic examples because of two main challenges: \emph{(i)} exactly solving the matching condition, and \emph{(ii)} applying the method beyond simple stabilization tasks.
First, most of the methods (see e.g., \cite{viola2007total,harandi2022solution,kotyczka2013local,acosta2005interconnection,acosta2009pdes}) aim to solve the so-called \emph{matching conditions} \emph{exactly and analytically}. The matching conditions are a set of nonlinear partial differential equations (PDEs), whose feasibility determines the possibility of shaping the desired closed-loop pH structure. However, the solution of the matching conditions is often prohibitively expensive or even impossible to achieve for complex control problems and systems. 
Second, while the broader motivation of energy-based methods is to realize \emph{desired behaviors} beyond mere stabilization, almost all IDA-PBC methods in the literature are limited to solving \emph{stabilization} tasks. This is mostly because the IDA-PBC technique has \emph{universal stabilizing} properties, i.e., it can be proven that \emph{all} asymptotically stabilizing controllers on control affine systems can be implemented through IDA-PBC designs \cite{ortega2002interconnection,ortega2004interconnection}. From a target pH system perspective, solving a stabilization task means that the closed-loop Hamiltonian shall possess a strict minimum at the desired equilibrium, leaving the other system variables as degrees of freedom to solve the matching conditions, rather than optimizing them for desired task-based behaviors.

The first mentioned problem is recently receiving attention, as IDA-PBC-inspired designs are being investigated together with machine learning approaches, aiming at using neural networks to approximate solutions of matching PDEs. In scenarios where exact matching is impossible to achieve, the use of IDA-PBC controllers and related closed-loop analysis under \emph{approximate matching conditions} would be highly beneficial to controller designs. In particular, \emph{physics-informed neural networks} \cite{raissi2019physics, cuomo2022scientific} have been applied to IDA-PBC \cite{sanchez2022stabilization,Sanchez-Escalonilla2024,plaza2022total}. These approaches consist of training neural networks to minimize cost functions encoding the residuals of the matching PDEs and closed-loop stability requirements, extending the applicability of IDA-PBC methods beyond examples in which matching PDEs can be solved analytically. Notably, in \cite{Sanchez-Escalonilla2024} a robustness analysis of the stability of the closed-loop equilibrium is performed using Lyapunov arguments (a strategy initially proposed in \cite{ortega2004interconnection}), providing robust theoretical guarantees on the closed-loop system emerging from the neural design. However, the application of neural network-based approaches, although potentially introducing flexibility in the task-based objective, are still limited to stabilization tasks. The second problem is rarely addressed. While attempts have been made to use IDA-PBC controllers for tasks other than stabilization (discussed in \cite[Sec.6]{ortega2004interconnection}), these efforts have resulted in isolated, yet notable works, such as \cite{yi2020orbital}, where the orbital stabilization problem is solved through exact matching.

In this paper, we address the two aforementioned limitations of IDA-PBC. In particular, we present a novel way to design IDA-PBC controllers capable of tackling general tasks, going beyond mere stabilization, while achieving, at the same time, approximate system matching along the optimal trajectories. This is a key aspect for extending the applicability of IDA-PBC designs beyond academic examples that rely on analytical solutions to the matching condition PDE. To do so, we cast IDA-PBC as a multi-objective optimal control problem. The cost function of the optimization problem combines two elements: \emph{(i)} a task-dependent performance metric used to identify a desired pH closed-loop structure and encode the desired energy-based behavior, and \emph{(ii)} a penalty term that accounts for the feasibility of achieving that closed-loop system using the IDA-PBC controller.

The multi-objective optimal control problem is solved using neural ODEs \cite{chen2018neural}. We \emph{(i)} parametrize the IDA-PBC controller, i.e., the desired-system matrices that represent the closed-loop pH dynamics, then \emph{(ii)} we numerically solve the neural ODE forward and evaluate the cost of the trajectory, and eventually \emph{(iii)} we optimize the parameters of the controller to minimize the cost through adjoint or backpropagation techniques. Instead of using a black-box neural ODE to parametrize our controller, we rely on dictionary learning \cite{tovsic2011dictionary} to obtain a sparse and interpretable expressions of the resulting control law and the closed-loop system. We test our approach using the electromechanical system introduced in \cite{ortega2004interconnection}) in the task of \emph{(i)} regulating the position of the capacitor plate and \emph{(ii)} discovering periodic oscillations of the capacitor plate.
In addition, the resulting analytical expression from the sparse dictionary learning enables the computation of the state transition matrix, a fundamental tool in stability analysis of orbits.

\section{Background}

\subsection{Port-Hamiltonian systems and IDA-PBC}

Port-Hamiltonian (pH) systems \cite{Duindam2009ModelingSystems,van2000l2,Rashad2020TwentyReview} are a class of control affine systems in the form:
\begin{equation}
\label{eq:pH}
    \begin{cases}
      \dot{x}= (J(x)-R(x))\partial_x H(x)+g(x)u \\
      y=g(x)^{\top}\partial_x H(x)
    \end{cases} \, 
\end{equation}
where $x\in \mathcal{D}\subseteq \mathbb R^n$ is the state, $u\in \mathcal{U}\subseteq \mathbb R^m$ is the input, $g(x)$ is input matrix, $J(x)=-J(x)^{\top}$ and $R(x)=R(x)^{\top}\geq 0$ are respectively skew-symmetric and positive semi-definite symmetric matrices representing the power-preserving and the dissipative components of the system. The non-negative function $H: \mathcal{D}\to \mathbb{R}^{+}$, called the \emph{Hamiltonian}, maps the state into the total energy of the system. $\partial_x H(x) \in \mathbb R^{n}$ denotes the gradient of $H$, represented as a column, and $\partial_x^{\top}H(x)$ denotes its transposed.
PH systems are \emph{passive} \cite{van2000l2} with respect to the Hamiltonian as a storage function:
\begin{equation}
\label{eq:pow}
    \dot{H}=-\partial_x H ^{\top}R(x)\partial_x H+y^{\top}u\leq y^{\top}u.
\end{equation}

In extreme synthesis, the motivation underlying pH methods in control are based on assessing desired energetic behaviors from the system matrices in (\ref{eq:pH}) and the power balance (\ref{eq:pow}), and using the Hamiltonian as a Lyapunov function for stability analysis \cite{Ortega2001PuttingControl}. 
Interconnection and damping assignment passivity-based control (IDA-PBC) \cite{ortega2002interconnection, ortega2004interconnection} is a technique that regulates the behavior of nonlinear systems assigning a pH structure to the closed-loop system, whose peculiar parametrization can be conveniently used to represent a desired target behavior. The goal of an IDA-PBC controller is to transform the nonlinear plant system
\begin{equation}
\label{eq:pH_affine}
      \dot{x}= f(x)+g(x)u 
\end{equation}
into the desired target pH system
\begin{equation}
\label{eq:pHdes}
      \dot{x}= (J_d(x)-R_d(x))\partial_x H_d(x). 
\end{equation}
Using the state feedback
$u=\beta(x)$, the desired closed loop system (\ref{eq:pHdes}) is obtained if 
\begin{equation}
\label{eq:closed-loopsubstitution}
    f(x) +g(x)\beta(x)=(J_d(x)-R_d(x))\partial_x H_d(x).
\end{equation}

While in the space spanned by the input matrix $g(x)$, (\ref{eq:closed-loopsubstitution}) can be satisfied directly by means of a proper choice of the feedback law $\beta(x)$, the difficulty arises in satisfying (\ref{eq:closed-loopsubstitution}) in the \emph{non actuated} coordinates. This can be explicitly seen by pre-multiplying the latter equation by the full rank left annihilator of $g(x)$, namely the full-rank matrix $g^{\perp}(x)\in \mathbb{R}^{(n-m)\times n }$ such that $g^{\perp}(x)g(x)=0$. The following $(n-m)$ scalar equations form a nonlinear PDE in the state variables, called \textit{matching conditions}:
\begin{equation}
\label{eq:matchingconditions}
    g^{\perp}(x)f(x) =g^{\perp}(x)(J_d(x)-R_d(x))\partial_x H_d(x).
\end{equation}
By solving the matching conditions (\ref{eq:matchingconditions}) for $g^{\perp}, J_d, R_d, H_d$, one achieves the target closed-loop system (\ref{eq:pHdes}), \emph{de facto} solving the IDA-PBC problem, using the state feedback:
\begin{equation}
\label{eq:statefeedback}
u=g^{+}(x)\{(J_d(x)-R_d(x))\partial_x H_d(x)-f(x)\}\, ,
\end{equation}
where $g^{+}(x)$ is the left pseudo-inverse of $g$.

\subsection{Neural Ordinary Differential Equations}
Neural ordinary differential equations (neural ODEs) \cite{chen2018neural} are a class of deep learning models in which the evolution of the hidden states is modeled as a continuous-time dynamical system, defined by an ODE whose dynamics are parameterized by a neural network.
Traditional feedforward neural networks apply a finite amount of transformations to the input data determined by the number of layers:
\begin{equation*}
    x_{i+1} =  f_i(x_i)\, ,
\end{equation*}
where $f_i$ indicates the $i^{\text{th}}$ layer of the neural network, and $x_{i}, x_{i+1}$ the input and output of layer $i$, respectively. Conversely, in neural ODEs, the hidden state evolves continuously over time according to an ODE:
\begin{equation*}
    \frac{dx}{dt} = f_\theta(x(t), t)\, ,
\end{equation*}
where the vector field $f_\theta$ is parametrized by a neural network of parameter $\theta$. To obtain the output, we can solve an initial value problem, i.e., we integrate the neural ODE from an initial state $x(t_0)$ at time $t_0$ to a final time $t_f$ using an ODE solver:
\begin{equation*}
\begin{split}
        x(t_f) &= \texttt{ODESolve}(f_\theta, x(t_0), t_0, t_f) \\ &=  x(t_0) + \int_{t_0}^{t_f}f_\theta(x(t), t) dt.
\end{split}
\end{equation*}

Similarly to traditional neural networks, we can optimize the parameters $\theta$ of the neural ODE by minimizing a loss function. In the case of a scalar-valued loss function $\mathcal{L}$, for example the mean-squared error, we obtain:
\begin{equation*}
    \mathcal{L}(\theta) = ||x_{\text{ref}} - \texttt{ODESolve}(f_\theta, x(t_0), t_0, t_f)||_2^2\, ,
\end{equation*}
where $x_{\text{ref}}$ is the reference output in the context of a regression task. This operation requires the calculation of the gradients of the loss with respect to $\theta$. This operation is known as the adjoint sensitivity method \cite{pontryagin2018mathematical}.
The training of neural ODEs is equivalent to the optimal control problem \cite{pontryagin2018mathematical}:
\begin{equation*}
    \begin{split}
        &\min_\theta \mathcal{L}(\theta)\\
        \text{subject to } &\frac{dx}{dt} = f_\theta(x(t), t) \hspace{10pt} \text{with } x_0 = x(t_0). \\
    \end{split}
\end{equation*}

\subsection{Sparse Dictionary Learning}\label{subsec:sparse_dictionary_learning}
Dictionary-learning methods are data-driven methods approximating nonlinear functions as
linear combinations of candidate dictionary functions, e.g., polynomials of degree $d$ or trigonometric functions \cite{tovsic2011dictionary}. Sparse dictionary-learning techniques additionally enforce sparsity to balance dictionary complexity with approximation accuracy, i.e., by identifying the smallest number of non-zero dictionary functions that can best approximate the nonlinear function. In the context of learning and controlling dynamical systems from data, the sparse identification of nonlinear dynamics method (SINDy) \cite{brunton2016discovering} can discover governing equations by relying on sparse dictionary learning.
In particular, given a set of $N$ input data $X = \{x_1, \cdots, x_N\}$ and labeled outputs $Y = \{y_1, \cdots, y_N\}$, we want to identify the unknown function $f:\mathcal{X}\rightarrow\mathcal{Y}$ that maps $x$ to $y$ as $y = f(x)$. To find the best approximation of $f$, we construct a dictionary of candidate functions $\Theta(X)= \{\theta_1(X), \cdots, \theta_D(X)\}$. Given this dictionary, we can write the input-output relation as:
\begin{equation*}
    Y = \Theta(X)\Xi,
\end{equation*}
where $\Xi$ is the matrix of learnable coefficients of coefficient to be found. Unlike standard neural networks, which impose no restriction on the function class being approximated, the choice of a dictionary restricts the set of possible functions. This restriction, however, reduces the number of learnable parameters to only the coefficients of the matrix $\Xi$.

While typically the coefficient matrix $\Xi$ is sparsified by sequentially thresholded least squares, alternative approaches exist. One such example is the differentiable $L_0$ regularization method for sparsifying neural networks \cite{louizos2017learning}. This approach relaxes the discrete nature of L$_0$ to allow efficient gradient-based optimization and can be used for sparsifying the coefficient matrix $\Xi$. In the context of sparse dictionary learning, using the L$_0$ norm is beneficial in the context of SINDy \cite{zheng2018unified, champion2020unified}. Moreover, recent work shows that the L$_0$ loss provides improved performance when combining dimensionality reduction using variational autoencoders \cite{kingma2013auto} with SINDy for discovering governing equations of stochastic dynamical systems~\cite{jacobs2023hypersindy}. Eventually, L$_0$-sparse dictionary models are used in \cite{botteghi2024parametric} to learn interpretable polynomial policies in the context of control of parametric partial differential equations with reinforcement learning.
We refer the reader to Appendix \ref{app:derivationL0} for the derivation of the L$_0$ regularization.

\section{Methodology}
In this section, we propose a general methodology for IDA-PBC that relies on neural ODEs optimization and sparse dictionary learning to learn controllers for nontrivial tasks in a principled and interpretable way.

\subsection{A multi-objective perspective on IDA-PBC}

Although not the most common perspective in the literature, IDA-PBC presents a \emph{multi-objective} nature. On one side, the shaping of the desired closed-loop system \eqref{eq:pHdes} should be driven by high-level task metrics, a fact motivated very deeply in all pH literature, i.e., \eqref{eq:pHdes} encodes a desired \emph{behavior} of the controlled system, beyond mere stability properties \cite{Ortega2001PuttingControl}. On the other hand, the structure of the plant \eqref{eq:pH} influences the achievable closed-loop systems, a constraint represented by the matching condition (\ref{eq:matchingconditions}).
In case of a stabilization task, the closed-loop Hamiltonian $H_d$, which acts as a Lyapunov function for the closed-loop system, should present a minimum at the desired equilibrium $x^*$. Thus, the IDA-PBC design focuses on solving (\ref{eq:matchingconditions}) in the remaining degrees of freedom in (\ref{eq:pHdes}). However, for a generic task, going beyond stabilization, the shape of $H_d$ may not be known a priori.
In this work, we formally present the IDA-PBC methodology in the form of a multi-objective optimization, aiming to find the best trade-off between achieving a desired behavior and solving the matching conditions. 
The decision variables are $\Psi=\{ J_d,R_d,H_d\}$\footnote{In principle, it is also possible to include $g^{\perp}$ as an additional degree of freedom.}. The cost function comprises a \emph{(i)} task-based term $J_{\text{task}}$, and \emph{(ii)} a cost penalizing a deviation from the matching condition (\ref{eq:matchingconditions}): \[J_{\text{mc}}=||g^{\perp}(x)(f(x)-f_d(x))||\, ,\] 
where $f_d(x)=((J_d(x)-R_d(x))\partial_x H_d(x))$ depends on the current guess for the decision variables $\Psi$. In this case the matching condition is evaluated on a $(n-m)$-dimensional vector representing the non-actuated coordinates, and the control law (\ref{eq:statefeedback}) is determined by the minimizer of the optimization $\Psi^*=\{ J_d^*,R_d^*,H_d^*\}=\textrm{argmin} (J_{\text{task}}+\lambda J_{\text{mc}})$. Denoting $f^*_d(x)=((J_d^*(x)-R_d^*(x))\partial_x H_d^*(x))$, the closed-loop system results in
\begin{equation}
      \dot{x}= f(x)+g(x)g^{+}(x)((f^*_d(x) - f(x))\, ,
\end{equation}
which can be written as:
\begin{equation}
      \dot{x}= f_d^*(x)-\eta(x)\, ,
\end{equation}
where
\begin{equation}
    \eta(x)=(I_n-gg^{+})(f_d^*-f).
\end{equation}

This form suggests another way to represent the matching conditions: \[J_{\text{mc}}=||\eta(x)||.\] In this form $g^{\perp}$ does not appear  and the matching condition is evaluated on the complete $n$-dimensional state space, in a form which implicitly assumes the use of the controller (\ref{eq:statefeedback}).

\subsection{Sparse Neural ODE for IDA-PBC}\label{subsec:sparseNODE_IDA}

\begin{figure}[h!]
    \centering
    \includegraphics[width=0.65\linewidth]{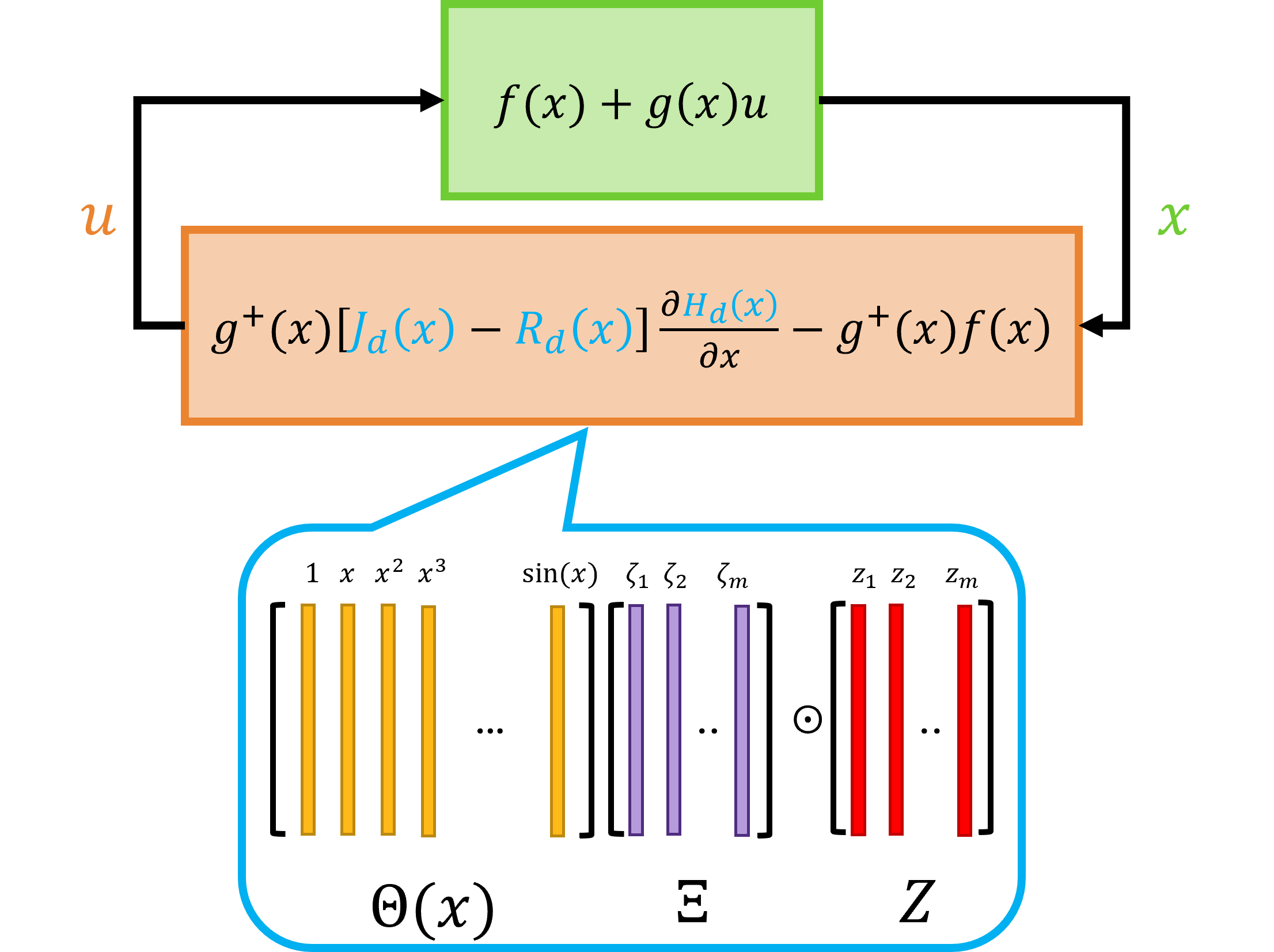}
    \caption{State-feedback control law relying on a parametrization of $J_d(x)$, $R_d(x)$, and $H_d(x)$ using differentiable sparse dictionary models.}
    \label{fig:figure_1}
\end{figure}

We aim to solve the following optimization problem:
\begin{equation}
    \begin{split}
        \min_{\Psi} \ \ \ &J_{\text{task}}(x,u) + \gamma J_{\text{mc}}(x) \\
        \text{s.t. } \ \ \ &\dot{x} = \big(J_d(x)-R_d(x)\big)\frac{\partial H_d(x)}{\partial x}-\eta(x), \hspace{10pt} x(0) = x_0. \\
    \end{split}
    \label{eq:approx_optimal_control_problem}
\end{equation}
The control variables are $\Psi = \{J_d, R_d, H_d\}$. To account for non-perfect matching, we include the residual $\eta(x)$ in the closed-loop system dynamics, i.e., the dynamical system in the constraint is the system obtained using the IDA-PBC controller (\ref{eq:statefeedback}) with current guesses of $J_d, R_d$, and $H_d$, which do not satisfy the matching conditions exactly, unless $\eta(x)=0$. We graphically visualize the state-feedback control law in Figure \ref{fig:figure_1}.

For complex tasks, the structure of $J_d, R_d, H_d$ may not be known a priori. While it is possible to parametrize $J_d, R_d, H_d$ with neural networks \cite{sanchez2022stabilization, Sanchez-Escalonilla2024, plaza2022total}, here we argue that a neural network does not allow for any type of a-posteriori interpretability. Therefore, following the work in \cite{botteghi2024parametric}, we rely on a completely differentiable sparse dictionary learning approach to approximate $J_d, R_d, H_d$ that we can use to solve our neural ODE optimization problem. We indicate the parametrization of the matrices as:
\begin{equation*}
    \begin{split}
        J_d(x) &= J_d(x;\Xi_{J_d},Z_{J_d}) \\
        R_d(x) &= R_d(x;\Xi_{R_d},Z_{R_d}) \\
        H_d(x) &= H_d(x;\Xi_{H_d},Z_{H_d}) \\
    \end{split}\, ,
    \label{eq:param_JdRdHd}
\end{equation*}
where $\Xi$ indicates the matrix of learnable coefficients and $Z$ the sparsification mask. In addition, we assume $J_d, R_d$ having the following structure:
\begin{equation}
\begin{aligned}
    J_d(x;\Xi_{J_d},Z_{J_d}) = 
        \begin{bmatrix}
            0 & a & \beta\\
            -a & 0 & c\\
            -\beta & -c& 0 
        \end{bmatrix}\, ,
    \end{aligned}
    \label{eq:jd}
\end{equation}
\begin{equation}
    \begin{aligned}
    R_d(x;\Xi_{R_d},Z_{R_d}) = 
        \begin{bmatrix}
            d & 0 & 0\\
            0 & e & 0\\
            0 & 0 & f 
        \end{bmatrix}\, ,
    \end{aligned}
    \label{eq:rd}
\end{equation}
where $a=a(x;\Xi_{J_d},Z_{J_d})$, $\beta=\beta(x;\Xi_{J_d},Z_{J_d})$, $c=c(x;\Xi_{J_d},Z_{J_d})$, $d=d(x;\Xi_{R_d},Z_{R_d})$, $e=e(x;\Xi_{R_d},Z_{R_d})$, and $f=f(x;\Xi_{R_d},Z_{R_d})$
are parametrized state-dependent functions. The differentiable parametrization of $J_d, R_d, H_d$ allows us to solve the optimization problem in \eqref{eq:approx_optimal_control_problem_par} using gradient-descent methods efficiently. In particular, we rely on the ADAM optimizer to solve:
\begin{equation}
    \begin{split}
        \min_{\Psi} \ \ \ &J_{\text{task}}(x,u) + \lambda J_{\text{mc}}(x) + \gamma J_{\text{sparse}}\\
        \text{s.t. } \ \ \ &\dot{x} = \big(J_d(x)-R_d(x)\big)\frac{\partial H_d(x)}{\partial x}-\eta(x), \hspace{10pt} x(0) = x_0, \\
    \end{split}
    \label{eq:approx_optimal_control_problem_par}
\end{equation}
where $J_{\text{sparse}}$ indicates the $L_0$ regularization loss introduced in Section \ref{subsec:sparse_dictionary_learning}, used to learn a sparse controller, specifically one with the least number of active terms.

\section{Numerical Results}

To test our approach, we study two control problems, namely \emph{(i)} regulation and \emph{(ii)} discovery of an oscillatory behavior, of an electrostatic microactuator, modeled as voltage-controlled capacitor with moving plate \cite{ortega2004interconnection}. The state is $x = (q, p, Q)$, the dynamic equations in pH form are:
\begin{equation}\label{eq:electromech}
    \begin{aligned}
        \begin{bmatrix}
            \dot q\\
            \dot p\\
            \dot{Q}
        \end{bmatrix}&=\Bigg(
        \begin{bmatrix}
            0 & 1 & 0\\
            -1 & 0 & 0\\
            0 & 0 & 0
        \end{bmatrix}-
        \begin{bmatrix}
            0 & 0 & 0\\
            0 & b & 0\\
            0 & 0 & 1/R
        \end{bmatrix}\Bigg)
        \begin{bmatrix}
            \frac{\partial H}{\partial q}\\
            \frac{\partial H}{\partial p}\\
            \frac{\partial H}{\partial Q}
        \end{bmatrix} +
        \begin{bmatrix}
            0\\
            0\\
            1/R \end{bmatrix}u,
            \end{aligned}\
\end{equation}
with $H(q,p,Q)=\frac{1}{2} k (q-q_0)^2+\frac{1}{2m}p^2+\frac{q}{2A\epsilon}Q^2$. Here $q$ is the airgap in the capacitor (position of the movable plate, with the position of the fixed plate at zero), $q_0$ the rest length of the spring attached to the moving plate, $p$ the momentum of the movable plate, $Q$ the charge of the capacitor, $A$ the area of the plate, $m$ the mass of the plate, $\epsilon$ the permittivity in the gap, $k$ and $b$ spring and damping coefficients, $R$ electric resistance, and $u$ input voltage. 

\subsection{Regulation of the Capacitor's Moving Plate}
The goal is to find a controller that can steer the plate of the capacitor to a desired reference $q^*$. We define our task-dependent cost function $J_{\text{task}}$ as:
\begin{equation}
    J_{\text{task}}(x, u) = J_{\text{reg}}(x) + \gamma_1 J_{\text{eff}}(u) \, ,
    \label{eq:Jtask_regulation}
\end{equation}
with
\begin{equation}
    \begin{split}
        J_{\text{reg}}(x) &= \int_0^T (q(t)-q^*)^2 dt \\
        J_{\text{eff}}(u) &= \alpha_3 \int_0^Tu(t)^2dt
        \, .
    \end{split}
\end{equation}

\begin{figure}
    \centering
    \includegraphics[width=0.65\linewidth]{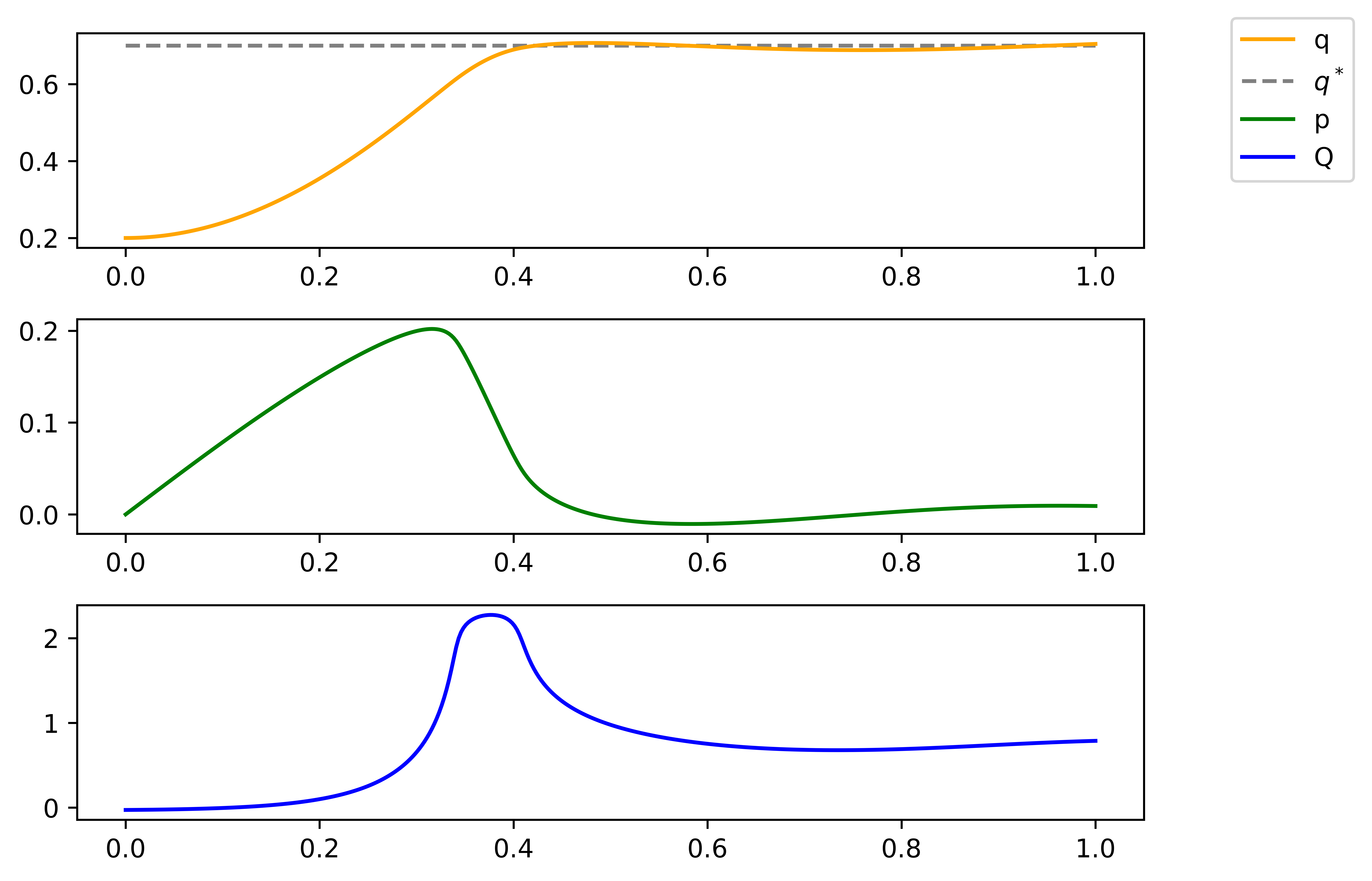}
    \caption{Evolution of the state variables.}
    \label{fig:placeholder}
\end{figure}

Using the task-dependent cost function defined in \eqref{eq:Jtask_regulation}, we solve the optimization problem in \eqref{eq:approx_optimal_control_problem} and we show the evolution of the state variables of the closed-loop system in Figure~\ref{fig:placeholder}.
Our approach is not only capable of solving the regulation task, but in addition it provides closed-form expressions of $H_d, J_d$ and $R_d$ (and consequently of the matching-condition residuals). In particular, for the regulation case we obtain:
\begin{equation}
\begin{split}
    H_d &= 6.1907 + 2.5271q - 0.4659p +4.090q^2 \\ &- 2.2953Q^2 -1.2348p^2Q\\
    a &= 0\\
    \beta &= -(0.2141  - 0.7414p)Q - 4.2964qp\\
    c &= 0.2217 + 1.7312pQ\\
    d &=  -0.1824 + 0.3306Q\\
    e &= 1.6327\\
    f &= 3.7035p +  6.4384qp +  6.4919pQ +  5.8333pQ^2 \\ &-1.6864qQ^3 + 4.6417pQ^3 \, ,\\
\end{split}
\end{equation}
where $a, \beta, c, d, e, f$ are the elements of $J_d$ and $R_d$, respectively, as shown in Equation \eqref{eq:jd} and \eqref{eq:rd}.
In addition, we report the learned value for $Q(0)$:
\begin{equation}
    Q(0)=-0.0275.
\end{equation}

\subsection{Discovery of an Oscillatory Behavior of the Capacitor's Moving Plate} \label{subsec:discovery_orbits}

Differently from the regulation example, we propose to solve a more challenging task, namely learning $J_d$, $R_d$, and $H_d$ such that the capacitor's moving plate oscillates according to the high-level metric. 
To learn such oscillations, we propose a variant of the cost function introduced in \cite{wotte2023discovering} to learn energy-efficient oscillations for a purely mechanical system. In particular, we include an additional term $J_{\text{period}}(x)$ to promote the periodicity of the state variable $Q$ that we call $J_{\text{period}}(x)$:
\begin{equation}
    J_{\text{task}}(x,u) = J_{\text{mid}}(x) + \
    \gamma_1 J_{\text{eigen}}(x) + \gamma_2J_{\text{eff}}(u) + \gamma_3 J_{\text{period}}(x),
    \label{eq:num_results_losses}
\end{equation}
with:
\begin{equation}
    \begin{split}
        J_{\text{mid}}(x) &= \frac{\alpha_1}{2} (q\Big(\frac{T}{2}\Big)-q^*)^2 \\
        J_{\text{eigen}}(x) &= \lambda_1\big(||q(t) - q(T-t)||_{\infty,T}  \\ 
        &+ \alpha_2||p(t)-p(T-t) ||_{\infty,T}\big) \\
        &+\frac{\lambda_2}{2}||p\Big(\frac{T}{2}\Big) ||_2^2 \\ 
        J_{\text{eff}}(u) &= \alpha_3 \int_0^T u(t)^2 dt \,, \\
        J_{\text{period}}(x) &= \alpha_4 |Q(0) - Q(T)|\, , 
    \end{split}
\end{equation}
where $q^*$ is a desired position to be reached at time $t=\frac{T}{2}$, $|| \cdot ||_2$ is the 2-norm, $T$ is the period of the oscillation, $|| \cdot ||_{\infty, T}=\max_{t\in\frac{T}{2}}(||\cdot||_1)$ with $|| \cdot ||_1$ indicates the 1-norm, and $\alpha_1,\alpha_2,\alpha_3,\alpha_4,\lambda_1, \lambda_2, \lambda_3$ are scaling factors balancing the contribution of each term. The cost function $J_{\text{task}}$ is composed of four main terms, namely $J_{\text{mid}}$, $J_{\text{eigen}}$,  $J_{\text{eff}}(u)$, and $J_{\text{period}}(x)$. The first term (weakly) enforces that $q(T/2)$ to be close to a desired value $q^*$. The second term is derived from the eigenmanifold theory \cite{wotte2023discovering} and promotes symmetric trajectories for $q$ and $p$ with the additional constraint on $p(T/2)$ to be equal to zero. The third term considers the minimization of the control effort and penalizes aggressive and energy-inefficient controllers, while the forth term enforces $Q(0)$ to be close to $Q(T)$. We utilize the loss function in Equation \eqref{eq:num_results_losses} to optimize the free parameters of $J_d(x), R_d(x), H_d(x)$ and the initial value of the variable $Q$, i.e., $Q(0)$ by solving the neural ODE optimization problem in \eqref{eq:approx_optimal_control_problem_par}.

\begin{figure}[h!]
    \centering
    \includegraphics[width=0.65\linewidth]{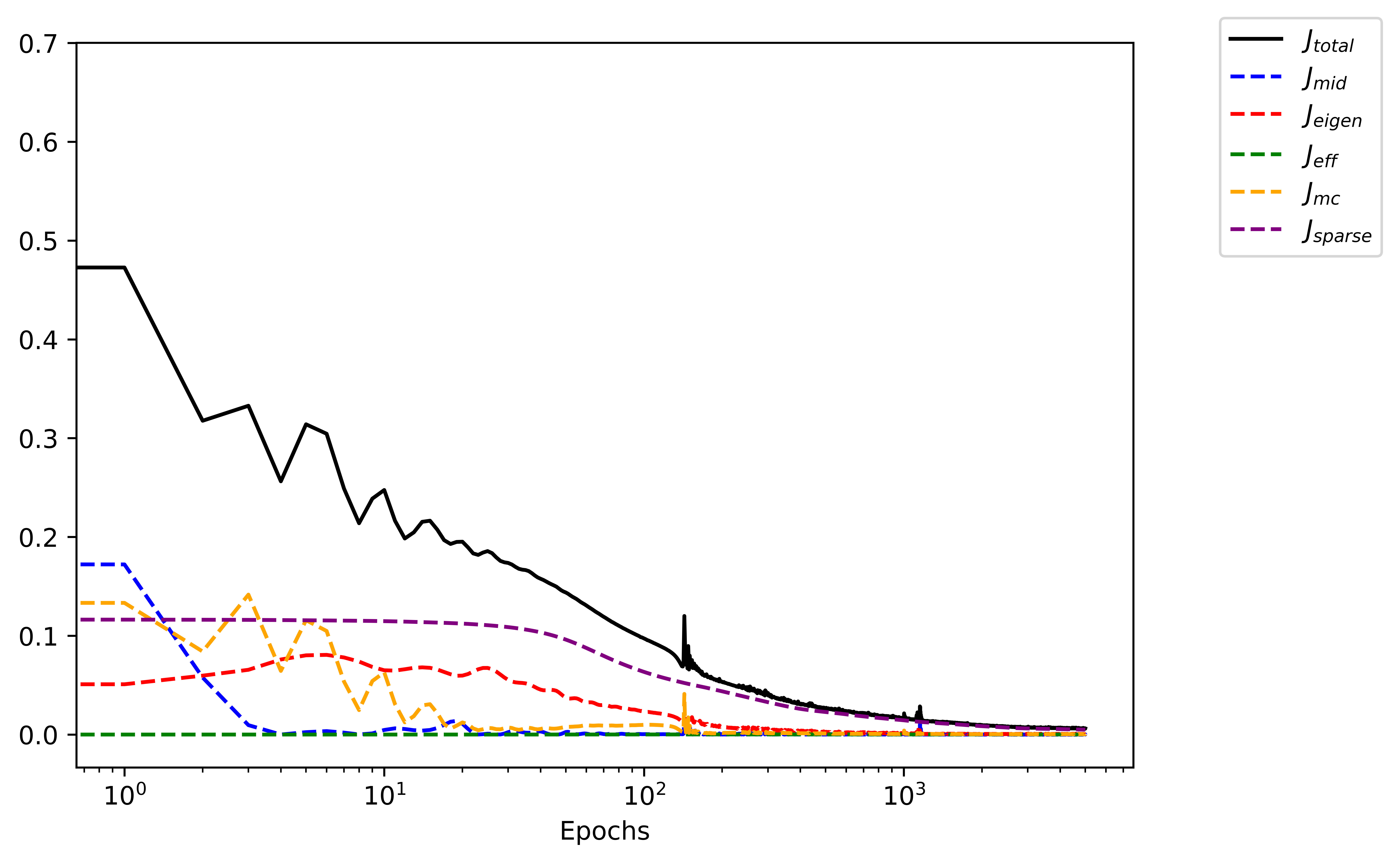}
    \caption{Trend of the loss functions over training.}
    \label{fig:losses}
\end{figure}

\begin{figure}[h!]
    \centering
    \includegraphics[width=0.65\linewidth]{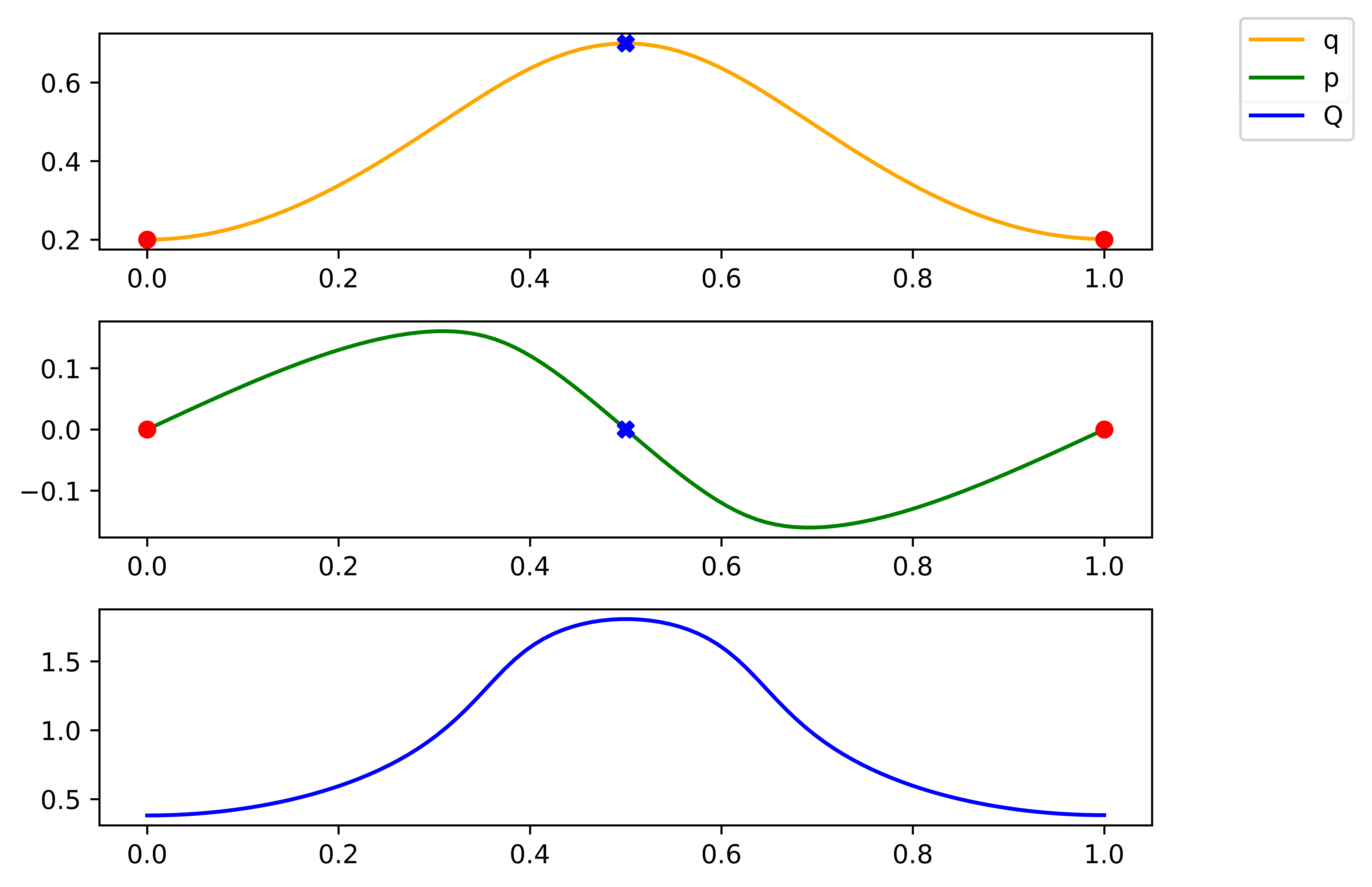}
    \caption{Evolution of the state variables for one period of oscillation $T$.}
    \label{fig:traj_electromech}
\end{figure}

Figure \ref{fig:losses} depicts the training losses over the 5000 epochs of training. The losses decrease over training and reach their minimum in the latest epochs of training. Despite the challenging optimization problem, our sparse neural ODE, built upon a library of polynomials of order 4, is sufficiently expressive to minimize the loss function.
In Figure \ref{fig:traj_electromech} we show the learned oscillatory behavior achieved by numerically solving the closed-loop system with the control input, defined as in Equation \eqref{eq:statefeedback}. We highlight with red dots the initial and final values for $q$ and $p$ and with blue crosses the desired target values for $T/2$, with period of oscillation equal to $1$ second. While we are not analytically solving the matching conditions, as it is very challenging for complex systems and tasks, we are capable of keeping the norm residual of the matching conditions $||\eta||_2^2$ small around the optimal trajectory with an average value of $0.0009$ over the first oscillation period $T$. 
Moreover, we report the closed-form expression for $H_d$, $J_d$, and $R_d$ that can be used for interpretability and further analysis. In particular, we obtain:
\begin{equation}
\begin{split}
    H_d &= 2.0114 + 2.2373q - 1.7219Q^2\\
    J_d &=
    \begin{bmatrix}
            0 & -0.3035 & -4.8265p \\
            0.3035 & 0 & 0.1458Q^4\\
            -4.8265p &  -0.1458Q^4 & 0
    \end{bmatrix} \\
    R_d &= 
    \begin{bmatrix}
            -0.0175 & 0 & 0\\
            0 & 0 & 0\\
            0 & 0 & (1.9354 + 1.5700Q + 2.6368Q^2 + 3.5828Q^3)p
    \end{bmatrix}. \\
\end{split}
\end{equation}
In addition, we report the learned value for $Q(0)$:
\begin{equation}
    Q(0)=0.3835.
\end{equation}

\begin{figure}[h!]
    \centering
    \includegraphics[width=0.65\linewidth]{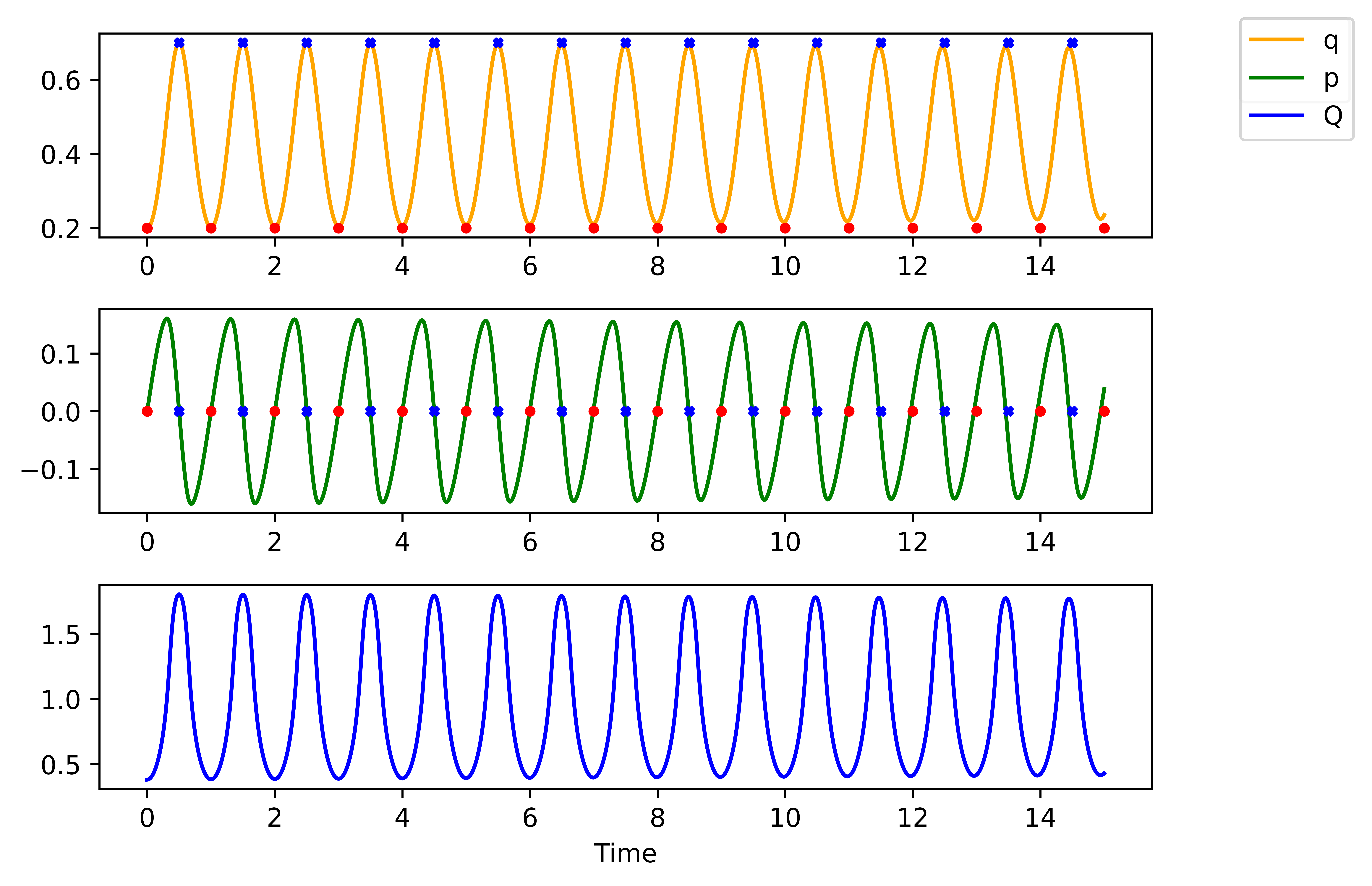}
    \caption{Evolution fo the state variables over multiple periods of oscillations.}
    \label{fig:elemech_traj_multiple}
\end{figure}

Eventually, we study the evolution of the systems over 15 periods of oscillations to show the effectiveness of our approach (Figure \ref{fig:elemech_traj_multiple}). Despite the training procedure only minimizing the loss functions over one single period of oscillation, the controller is capable of sustaining the periodic oscillations for multiple periods with just a noticeable deviation after 10 periods.

\subsubsection{Stability of the Learned Periodic Oscillation}

The sparse approximation of $J_d, R_d, H_d$ allows us to compute the state transition or system matrix (STM), which describes the linearized evolution of infinitesimally small perturbations. 
We evaluate the STM along a near-periodic solution of the system across one period $T$. This is known as the monodromy matrix $M$, and its eigenvalues describe the stability of a periodic orbit \cite{hartman2002ordinary}.

We take the flow $\phi(t,x_0) : x(t_0) \rightarrow x(t)$ of a dynamical system $ \dot{x} = f(x)$, and consider a small perturbation in initial condition $\delta\bar{x}_0$ from a reference trajectory $\bar{x}(t)$. The perturbation at time $t$ from the reference is given by
\begin{equation}
    \delta\bar{x}(t) = \phi(t; \bar{x}_0 + \delta\bar{x}_0) - \phi(t;\bar{x}_0).
\end{equation}
Since $\delta\bar{x}_0$ is small, we take the Jacobian of the flow map
\begin{equation}
     \delta\bar{x}(t) = \frac{\partial \phi(t;\bar{x}_0)}{\partial {x}_0} \delta \bar{x}_0,
     \label{eq:STM}
\end{equation}
which defines the state transition matrix 
\begin{equation}
    \Phi(t,t_0) = \frac{\partial \phi(t;\bar{x}_0)}{\partial x_0}\, ,
\end{equation}
describing the linearized evolution of infinitesimally small perturbations. Since we do not have an analytic expression for $\phi(t,x_0)$, we must find $\Phi(t,t_0)$ by numerical integration. We take the time derivative of \eqref{eq:STM}, and given that the time derivative of the flow is the system dynamics $f(x)$, we consider the system
\begin{equation}
    \dot{\Phi}(t,t_0) = \frac{\partial}{\partial x_0} \bigg( f(\bar{x}(t))\bigg) = \frac{\partial f}{\partial \bar{x}} \frac{\partial \bar{x}(t)}{\partial x_0}.
    \label{eq: STM sys}
\end{equation}
This creates the initial value problem
\begin{equation}
    \dot{\Phi}(t,t_0) = Df(\bar{x}(t)) \Phi(t,t_0), \quad \text{where } \Phi(t_0,t_0) = I\,,
\end{equation}
which must be numerically integrated with the system dynamics to find $\Phi(t,t_0)$ at a given time. 

The analytical nature of our method enables the computation of the monodromy matrix and, therefore, the characterization of the orbit's stability. For the controlled system \eqref{eq:approx_optimal_control_problem_par} with initial conditions $x = [0.2;	0;	0.3835]$, which correspond to a near periodic behaviour, this gives the monodromy matrix
\begin{equation}
    M = 
    \begin{bmatrix}
        0.9972 &    0.0233 &    0.0023\\
       -0.0615 &    0.9964 &    0.2547\\
       -0.0033 &    0.0315 &    1.0055\\
    \end{bmatrix}\, ,
\end{equation}
with eigenvalues 1.0812, 0.9990, 0.9188. For a truly periodic behaviour, we expect an eigenvalue of one, with the eigenvalue 0.9990 indicating a near-periodic behavior and a small instability associated with the eigenvalue 1.0812. The associated eigenvectors of $M$ correspond to the local manifold directions. Computing the STM permits the use of other mathematical tools such as differential correction methods and Floquet theory, which decomposes the STM into a periodic function $P(t)$ and an exponential term with constant matrix multiplier $R$ in the expression $\Phi(t) = P(t) e^{Rt}$ \cite{hartman2002ordinary}.

\section{Conclusion}
In this work, we propose a numerical approach for tackling regulation and periodic-orbit stabilization taska using IDA-PBC methodology without solving the matching PDEs exactly. In particular, we parametrize the desired closed-loop system, namely $H_d, J_d, R_d$ using sparse dictionary models. We optimize the learnable parameters of the neural ODE by minimizing a task-dependent cost and the residuals of the matching conditions. 
Our numerical results show that our approach is capable of shaping the closed-loop system to solve complex tasks, spanning from regulation to discovery of periodic behaviours, while keeping the residuals of the matching conditions minimal along the optimal trajectories. In addition, the use of sparse dictionary learning allows to obtain closed-form expressions of the learned controller that can be used for stability analysis through computing the state transition matrix.

\section*{Acknowledgments}
NB acknowledges the Project “Reduced Order Modeling and Deep Learning for the real-time approximation of PDEs (DREAM)” (Starting Grant No. FIS00003154), funded by the Italian Science Fund (FIS) - Ministero dell'Università e della Ricerca.

\bibliographystyle{unsrt}  
\bibliography{references}  

\appendix

\section{Differentiable L$_0$ Regularization}\label{app:derivationL0}
Let $d$ be a continuous random variable distributed according to a distribution $p(d| \psi)$, where $\psi$ indicates the parameters of $p(d| \psi)$. Given a sample from $d \sim p(d|\psi)$, we can define:
\begin{equation*}
    z= \min(1, \max(0, d)).
    \label{eq:hard-sigmoid-rectification}
\end{equation*}
The hard-sigmoid rectification allows the gate to be exactly zero. Additionally, we can still compute the probability of the gate being action, i.e., non-zero, by utilizing the cumulation distribution function $P$:
\begin{equation*}
    p(z\neq 0| \psi) = 1 - P(d \leq 0|\psi).
\end{equation*}
We choose as candidate distribution a binary concrete \cite{maddison2016concrete, jang2016categorical}. 
We can now optimize the parameters $\psi$ of the distribution and introduce the L$_0$ regularization loss as:
\begin{equation*}
    L_0(\psi) = \sum_{j=1}^{|\xi|} (1-P_{\bar{d}_j}(0|\psi)) = \sum_{j=1}^{|\xi|} \sigma(\log \alpha_j - \beta \log \frac{\gamma}{\zeta}),
\label{eq:L0_loss}
\end{equation*}
where $\xi$ are the parameters of the model we want to sparsify and $\sigma$ corresponds to the sigmoid activation function. At test time, we can estimate the sparse parameters $\xi^0$ by:
\begin{equation*}
    \begin{split}
        z &= \min(1, \max(0, \sigma(\log \alpha)(\zeta - \gamma) + \gamma)) , \, \\
        \xi^0 &= \xi \odot z, \,
    \end{split}
\end{equation*}
where  $\odot$ corresponds to the element-wise product. For a complete derivation of the L$_0$, including code tutorials, we refer the reader to \cite{botteghi2024sparsifying}.

\end{document}